\documentclass{article}
\usepackage{spconf,amsmath,epsfig}
\usepackage{bm}
\usepackage{multirow}
\begin{document}\sloppy

\def\x{{\mathbf x}}
\def\L{{\cal L}}

\title{Self-weighted Multiview Metric Learning by Maximizing the Cross Correlations}
%
\name{Huibing Wang, Jinjia Peng and Xianping Fu}
\address{Information Science and Technology College, Dalian Maritime University, Dalian 116026, China}

\maketitle

\begin{abstract}
	With the development of multimedia time, one sample can always be described from multiple views which contain compatible and complementary information. Most algorithms cannot take information from multiple views into considerations and fail to achieve desirable performance in most situations. For many applications, such as image retrieval, face recognition, etc., an appropriate distance metric can better reflect the similarities between various samples. Therefore, how to construct a good distance metric learning methods which can deal with multiview data has been an important topic during the last decade.
	
	In this paper, we proposed a novel algorithm named Self-weighted Multiview Metric Learning (SM$^2$L) which can finish this task by maximizing the cross correlations between different views. Furthermore, because multiple views have different contributions to the learning procedure of SM$^2$L, we adopt a self-weighted learning framework to assign multiple views with different weights. Various experiments on benchmark datasets can verify the performance of our proposed method.
	
\end{abstract}
\section{Introduction}
\label{sec:intro}

With the coming of information period, massive data has been generated from various fields to meet the requirement of different people. It is a common phenomena that one same sample can always be described by different tools (or techniques). For example, an image can be represented by several descriptors, such as Local Binary Patterns (LBP) \cite{ojala2002multiresolution}, Histogram of Oriented Gradient (Hog) \cite{hu2013performance} and Locality-constrained Linear Coding (LLC) \cite{wang2010locality}. Because these descriptors contain different which is compatible and complementary, how to exploit the multiview data to further improve the performance of various applications \cite{wu2018deep,wu20193D,wu2018andwhere} is of vital importance but challenging. 

For many applications, such as image retrieval \cite{wang2017effective}, face recognition \cite{wang2016semantic}, etc. \cite{feng2018learning, wang2015effective}, learning an appropriate distance metric can help these applications to measure the similarities between samples precisely. Xing et al.\cite{xing2003distance} proposed a convex optimization function, which minimizes the distances between samples from the same class. Meanwhile, they introduced two constraints to obtain a better metric. Weinberger et al. \cite{weinberger2006distance} introduced a DML method named Large Margin Nearest Neighbor (LMNN), which refers the thoughts of SMVs \cite{joachims2006training}. Information-Theoretic Metric Learning (ITML) \cite{davis2007information} is another well-known DML methods which combimes information theory \cite{cover2012elements} with distance metric learning (DML) and achieves good performances. Neighborhood components analysis (NCA) \cite{goldberger2005neighbourhood} trains a distance metric by maximizes the probabilities of that samples and their stochastic neighbors are from the same class. Even though these algorhtms have wide influences on this field, most DML methods cannot deal with multiview data simultaneously and waste a lot of important information.

In order to deal with the problem above, multiview learning  \cite{wang2015robust,wang2018multiview} has been studied recently. Kumar et al. \cite{kumar2011co} developed a co-regularized framework to minimize the distinctons between all views. All views are forced to learn from each other via the proposed framework. Xia et al.\cite{xia2010multiview} extended spectral embedding into multiview mode and proposed a self-weighted method to calcuate the weights of all views. Canonical correlation analysis (CCA) \cite{hardoon2004canonical} is another good multiview method which is an extension of Principal component analysis (PCA) \cite{jolliffe2011principal} and constructs the low-dimensional subspace by maximizing the cross correlations of each two views. Zhai et al. \cite{zhai2012multiview} combined global consistency with local smoothness and obtained a multiview metric learning method. Futhermore, some dimension reduction techniques can also be utilized as DML methods, such as PCA \cite{jolliffe2011principal} and LDA \cite{mika1999fisher} .Till now, most distance metric leanring methdos cannot deal with multiview data and obtain better performances. It is essential for researchers to conduct more studies on this field \cite{Wang2016Iterative}.

In this paper, we propose a novel multiview metric learning methods named Self-weighted Multiview Metric Learning (SM$^2$L) which can fully exploit multiview data to measure similarities between differetn samples. The proposed SM$^2$L first introduced the maximum margin criterion for all views to maximize the distances between samples from different classes in each view. Then, in order to force all views to learn from each other, SM$^2$L aims to maximize the cross correlations between the multiview features of one same sample. Finally, because multiple views have different impacts on the construction process of distance metric, SM$^2$L equips multiple views with different weights which can be learned automatically. 

This paper is organized as follows: in section 2, we introduced some basic knowledge and some related works. In section 3, we introduced the construction procedure of our proposed SM$^2$L in detail and the solving procedure of it. In section 4, we conducted several experiments to show the excellent performance of our proposed method. And section 5 made a conclusion in detail. 

\section{Related works}

In this section, we first introduced some basic knowlege about multiview distance metric learning. Then, we introduced 2 related algorithms. 

\subsection{Basic Knowledge}

Assume we are given a set of multiview data $\bm{X} = \left\{ \bm{X}^v \in \Re^{D_v \times n}, v=1,\cdots,m \right\}$ which consists of $n$ samples from $m$ views, where $\bm{X}^v =\left\{\bm{x}^v_1,\bm{x}^v_2,\cdots,\bm{x}^v_n \right\}\in \Re^{D_v \times n}$ contains all features from the $v$th view. For any two features $\bm{x}^v_i$ and $\bm{x}^v_j$ in the $v$th view, their Mahalanobis distance can be computed as:

\begin{equation}
\label{eq1}
\begin{array}{l}
d_{\bm{A}^v}\left(\bm{x}^v_i,\bm{x}^v_j\right) = \sqrt{\left(\bm{x}^v_i-\bm{x}^v_j\right)^T \bm{A}^v\left(\bm{x}^v_i-\bm{x}^v_j\right)}
\end{array}
\end{equation}

Where $\bm{A}^v$ is the distance metric matrix for the $v$th view. $d_{\bm{A}^v}\left(\bm{x}^v_i,\bm{x}^v_j\right)$ represents the distance between $\bm{x}^v_i$ and $\bm{x}^v_j$. Because $\bm{A}^v$ is a semi-definite matrix, it can be be decomposed into $\bm{A}^v =\bm{W}^v\left(\bm{W}^v\right)^T$, $\bm{W}^v \in \Re^{d \times D_v}$ and $ d_v \le D_v$ using eigenvalue decomposition.

For multiview distance metric learning methods, it's essential for all views to learn from each other to improve the quality of distance metric for each view. And for any distance metric, it should satisfy these four properties as follows:

1) triangular inequality: $d_{\bm{A}^v}\left(\bm{x}^v_i,\bm{x}^v_j\right)+d_{\bm{A}^v}\left(\bm{x}^v_j,\bm{x}^v_k\right) \ge d_{\bm{A}^v}\left(\bm{x}^v_i,\bm{x}^v_k\right) $;

2) non-negativity: $d_{\bm{A}^v}\left(\bm{x}^v_i,\bm{x}^v_j\right) \ge 0$;

3) symmetry: $d_{\bm{A}^v}\left(\bm{x}^v_i,\bm{x}^v_j\right) = d_{\bm{A}^v}\left(\bm{x}^v_j,\bm{x}^v_i\right)$;

4) distinguishability: $d_{\bm{A}^v}\left(\bm{x}^v_i,\bm{x}^v_j\right) =0 \leftrightarrow \bm{x}^v_i = \bm{x}^v_j$;

\subsection{Related Algorithms}
In this section, we introduced some related algorithms which have been refered in our proposed SM$^2$L.

\subsubsection{Multiview Spectral Embedding}

Multiview Spectral Embedding (MSE) \cite{xia2010multiview}  is a well-known multi-view dimension reduction method which can assign all views with appropriate weights automatically. It encodes different features from multiple views to achieve a physically
meaningful embedding. Furthermore, MSE integrates laplacian graphs from multiple views via global coordinate alignment, which help all views to learn from each other. And the objective function of MSE has been shown as follows:

\begin{equation}
\label{eq2}
\begin{array}{l}
\mathop {\arg \min }\limits_{\bm{\alpha} ,\bm{Y}} \sum\limits_{v = 1}^m {\alpha _v
	tr\left( {\bm{Y}\bm{L}^{\left( v \right)}\bm{Y}^T} \right)} \\
s.t.\;\bm{Y}\bm{Y}^T = \bm{I};\;\sum\limits_{v = 1}^m {\alpha _v } = 1,\;\;\alpha _v \ge 0
\\
\end{array}
\end{equation}

$\bm{\alpha} = \left[ {\alpha _1 ,\alpha _2 , \cdots ,\alpha _m } \right]$ is a set of coefficients which can reflect the importance of different views. $\bm{L}^{\left( v \right)}$ is the laplacian graph for features in the $v$th view. It reflects the neighborhood relationship between features in the $v$th view.  And $\bm{Y}$ is the commen low-dimensional representation for the original multi-view data. In order to obtain the optimal $\bm{Y}$, MSE develops an iterative optimization procedure to update $\bm{\alpha}$ and $\bm{Y}$ alternately. However, MSE is not a multiview metric learning method which cannot measure similarities between multiview data. 

\subsubsection{Generalized Multi-view Analysis}

Sharma et al. \cite{sharma2012generalized} proposed a multiview learning method to construct subspace for the task of classification \cite{wu2019cycle}. They extended linear discriminate analysis (LDA) \cite{mika1999fisher} into multiview mode by maximizing the cross correlations between each two views as follows:

\begin{equation}
\label{eq3}
\begin{array}{l}
\mathop{\arg \max} \limits_{\bm{P}^1,\bm{P}^2,\cdots,\bm{P}^m} \sum\limits_{i=1}^m \alpha_i tr\left((\bm{P}^i)^T \bm{S}_b^i (\bm{P}^i) \right) \\
~~~~~~~~~~~~~~~~+ \sum \limits_{i<j}\lambda_ij tr\left((\bm{P}^i)^T \bm{X}^i(\bm{X}^j)^T\bm{P}^j \right)\\

~~~s.t. \sum \limits_{i=1}^m \gamma_i tr\left((\bm{P}^i)^T \bm{S}_w^i (\bm{P}^i) \right) =\bm{I}_p\\
\end{array}
\end{equation}

Where $\bm{P}_i$ represents the projection matrix for features in the $i$th view. $\bm{S}_b^i$ is the between-class scatter matrix while $\bm{S}_w^i$ is the with-in scatter one. $\alpha_i$ and $\gamma_i$ are the weights for between  and with-in class degree the $i$th view. And $\lambda_ij$ is the regularized parameter. Even though Eq.\ref{eq3} maximize the the cross correlations between each two views, they cannot assign all views with different weights automatically, which causes too many parameters need to be set during the experiments.

\section{The Proposed SM$^2$L}

\begin{figure*}[htbp]
	\label{fig1}
	\centering
	\includegraphics[width=.85\textwidth]{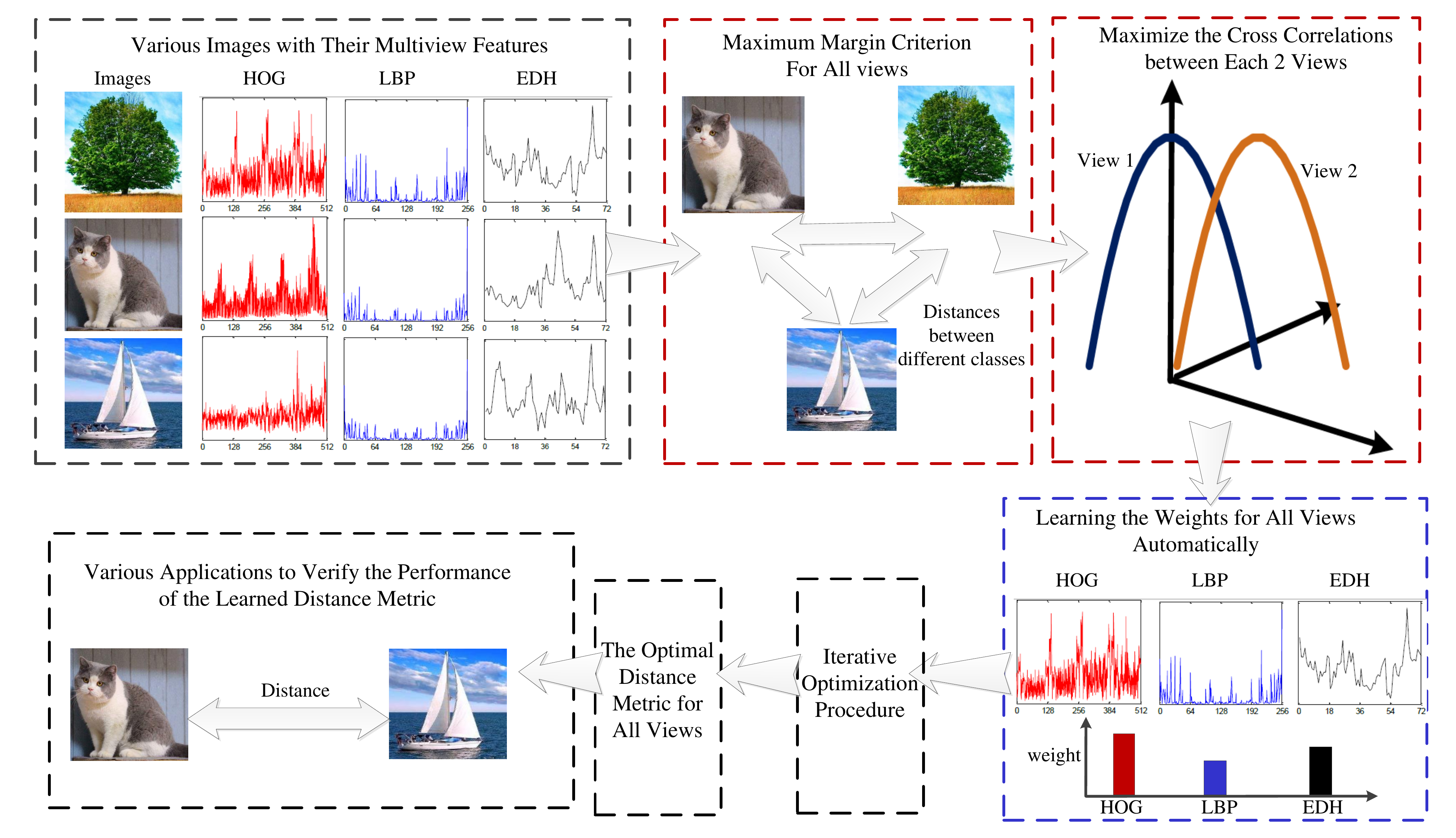}
	\caption{ The working procedure of SM$^2$L}
\end{figure*}

In this paper, we describe the construction procedure of SM$^2$L in detail as fig.\ref{fig1}. SM$^2$L first adopts maximize maximum margin criterion to help all views to construct distance metric matrix by maximize the distances between features from different classes. Then, SM$^2$L forces all views to learn from each other by maximizing the cross correlations between each two views. Finally, SM$^2$L assigns all views with different weights automatically to fully exploit the infromation of multiview data.

\subsection{The Construction Procedure of SM$^2$L }

For multiview data $\bm{X} = \left\{ \bm{X}^v \in \Re^{D_v \times n}, v=1,\cdots,m \right\}$ together with pairwise constraints $S$ and $D$ as follows:

$S: \forall (\bm{x}^v_i,\bm{x}^v_j) \in S$,     $~~~~~\bm{x}^v_i,\bm{x}^v_j \in $  same class

$D:\forall (\bm{x}^v_i,\bm{x}^v_j) \in D$,     $~~~~\bm{x}^v_i,\bm{x}^v_j \in $  different classes   

Then, according to the constraints above, SM$^2$L aims to maximize the distances between features from differetn classes while minimizing the distances between features in the same class as follows:

\begin{equation}
\label{eq4}
\begin{array}{l}

\mathop{\arg\max}\limits_{\bm{A}^1,\bm{A}^2,\cdots,\bm{A}^m} \sum\limits_{v=1}^m \left\{ \sum\limits_{i,j:(\bm{x}^v_i,\bm{x}^v_j)\in D} \frac{d^2_{\bm{A}^v}(\bm{x}^v_i,\bm{x}^v_j)}{N_D} \right.\\
~~~~~~~~~~~~~~~~~~~\left. -\sum\limits_{i,j:(\bm{x}^v_i,\bm{x}^v_j)\in S} \frac{d^2_{\bm{A}^v}(\bm{x}^v_i,\bm{x}^v_j)}{N_S}    
\right\}\\
\end{array}
\end{equation}

Where $N_S$ and $N_D$ are the numbers of pairwise constraints in $S$ and $D$. It is clearly that Eq.\ref{eq4} maximizes the distances between features from the different classes for all views. Due to Eq.\ref{eq1} and  $\bm{A}^v =\bm{W}^v\left(\bm{W}^v\right)^T$, Eq.\ref{eq4} can be further transformed as follows:

\begin{equation}
\label{eq5}
\begin{array}{l}
\mathop{\arg\max}\limits_{\bm{W}^1,\cdots,\bm{W}^m} 
\sum \limits_{v=1}^m

tr\left\{ (\bm{W}^v)^T \sum\limits_{(\bm{x}^v_i,\bm{x}^v_j)\in D} \frac{\left( \bm{x}^v_i-\bm{x}^v_j\right)\left( \bm{x}^v_i-\bm{x}^v_j\right)^T}{N_D} \bm{W}^v	\right.\\

~~~~~~~~~~~~~~~~\left. -(\bm{W}^v)^T \sum\limits_{(\bm{x}^v_q,\bm{x}^v_p)\in S} \frac{\left( \bm{x}^v_q-\bm{x}^v_p\right)\left( \bm{x}^v_q-\bm{x}^v_p\right)^T}{N_S} \bm{W}^v
\right\}

\end{array}
\end{equation}

And Eq.\ref{eq5} equals to 

\begin{equation}
\label{eq6}
\begin{array}{l}
\mathop{\arg\max}\limits_{\bm{W}^1,\bm{W}^2,\cdots,\bm{W}^m} 
\sum \limits_{v=1}^m tr\left\{(\bm{W}^v)^T \left( \bm{M}_D-\bm{M}_S  \right)\bm{W}^v \right\}
\\
s.t. (\bm{W}^v)^T\bm{W}^v = \bm{I}, v=1,2,\cdots,m
\end{array}
\end{equation}

Where $M_D = \sum\limits_{(\bm{x}^v_i,\bm{x}^v_j)\in D} \frac{\left( \bm{x}^v_i-\bm{x}^v_j\right)\left( \bm{x}^v_i-\bm{x}^v_j\right)^T}{N_D}$ and $M_S = \sum\limits_{(\bm{x}^v_i,\bm{x}^v_j)\in S} \frac{\left( \bm{x}^v_i-\bm{x}^v_j\right)\left( \bm{x}^v_i-\bm{x}^v_j\right)^T}{N_S}$. Therefore, the construction of $\bm{A}^v$ is equal to find the optimal $\bm{W}^v$.

However, Eq.\ref{eq5} cannot integrate infromation from multiple views. In order to fully take multiview data into consideration, SM$^2$L adopts the idea from \cite{sharma2012generalized} and maximizes the cross correlations betwee each 2 views as follows:

\begin{equation}
\label{eq7}
\begin{array}{l}
\mathop{\arg\max}\limits_{\bm{W}^1,\bm{W}^2,\cdots,\bm{W}^m} 
\sum \limits_{v=1}^m tr\left\{(\bm{W}^v)^T \left( \bm{M}_D-\bm{M}_S  \right)\bm{W}^v \right\} \\

~~~~~~ + \sum\limits_{v\neq w} \lambda_{ij} tr\left((\bm{W}^v)^T\bm{X}^v(\bm{X}^w)^T\bm{W}^w \right)
\\
s.t. (\bm{W}^v)^T\bm{W}^v = \bm{I}, v=1,2,\cdots,m
\end{array}
\end{equation}

Through Eq.\ref{eq7}, all views can learn from each other to further improve the performance of distance metric learning. However, because multiple views have different impacts on the construction of their distance metric matrix, it is essential for SM$^2$L to learn the weights of these views automatically. Therefore, the objective function of SM$^2$L can be organized as follows:

\begin{equation}
\label{eq8}
\begin{array}{l}
\max \mathcal{G}\left(\bm{\alpha},\bm{W}^1,\bm{W}^2,\cdots,\bm{W}^m \right)\\
~~~~~~= \sum \limits_{v=1}^m \alpha^r_v tr\left((\bm{W}^v)^T \left( \bm{M}_D-\bm{M}_S  \right)\bm{W}^v \right) \\

~~~~~~ + \sum\limits_{v\neq w} \frac{\alpha^r_v+\alpha^r_w}{2\eta} tr\left((\bm{W}^v)^T\bm{X}^v(\bm{X}^w)^T\bm{W}^w \right)
\\
s.t. (\bm{W}^v)^T\bm{W}^v = \bm{I}, v=1,2,\cdots,m \\
~~~~~~\sum\limits_{v=1}^m \alpha_v=1, r>1
\end{array}
\end{equation}

Where $\bm{\alpha}=\left[\alpha_1,\alpha_2,\cdots,\alpha_m \right]$ consists of weights of multiple views. $r>1$ ensures that all views have impacts on the construction of distance metric matrix. Therefore, once the optimal $\bm{W}^v$ is obtained for the $v$th view, the distance metric matrix can be calculated by $\bm{A}^v =\bm{W}^v\left(\bm{W}^v\right)^T$.

\subsection{The Solving Procedure of SM$^2$L}

In this section, we introduce the solving procedure of SM$^2$L in detail. Because it is difficult to obtain $\bm{\alpha}$ and $\bm{W}^1,\bm{W}^2,\cdots,\bm{W}^m$ simultaneously. SM$^2$L adopts alternately iteration to update the variables separately. 

\textbf{Update} $\bm{W}^1,\bm{W}^2,\cdots,\bm{W}^m$:  SM$^2$L keeps $\bm{\alpha}$ unchanged to update $\bm{W}^v,v=1,2,\cdots,m$ solely. It has been verified in \cite{sharma2012generalized} that the optimization procedure for  $\bm{W}^1,\bm{W}^2,\cdots,\bm{W}^m$ is a genelarized eigenvalue decomposition problem. The solving procedure of SM$^2$ is similar with that in \cite{sharma2012generalized}.

\textbf{Update} $\bm{\alpha}$: After $\bm{W}^1,\bm{W}^2,\cdots,\bm{W}^m$ are obtained, how to calculate $\bm{\alpha}$ is a main problem. By using a Lagrange multiplier $\lambda$ to take the constraint $\sum_{v=1}^m \alpha_v=1$ into consideration, we get the Lagrange function as:
\begin{equation}
\label{eq9}
\mathcal{L}\left(\bm{\alpha},\lambda\right) = \mathcal{G}\left(\bm{\alpha},\bm{W}^1,\bm{W}^2, \cdots,\bm{W}^m\right) - \lambda \left( \sum\limits_{v=1}^m \alpha_v-1 \right)
\end{equation}

By setting the derivative of $\mathcal{L}\left(\bm{\alpha},\lambda\right)$ with respect to $\alpha_v$ and $\lambda$ to zero, we can get:
\begin{equation}
\label{eq10}
\left\{ {\frac{\partial \mathcal{L}\left(\bm{\alpha},\lambda\right)}{\partial \alpha_v} = \frac{\partial \mathcal{G}\left(\bm{\alpha},\bm{W}^1,\bm{W}^2, \cdots,\bm{W}^m\right)}{\partial \alpha_v} -\lambda = 0, v=1,2,\cdots,m  \atop  \frac{\partial \mathcal{L}\left(\bm{\alpha},\lambda\right)}{\partial \lambda} = \sum\limits_{v=1}^m \alpha_v-1 =0 ~~~~~~~~~~~~~~~~~~~~~~~~~~~~~~~~~~~~~~~~~~~~~~~~}\right.
\end{equation} 

Through Eq.\ref{eq9}, we can get $\alpha_v$:

\begin{equation}
\label{eq11}
\alpha_v = \frac{\left(1/ tr\left( {\left( {\bm{U}^v} \right)^T\mathcal{J}^v \bm{U}^v} \right) \right)^{1/\left(r-1\right)}}{\sum\limits_{v=1}^m\left(1/ tr\left( {\left( {\bm{U}^v} \right)^T\mathcal{J}^v \bm{U}^v} \right) \right) ^{1/\left(r-1\right)}}
\end{equation}

where 

\begin{equation}
\label{eq12}
\begin{array}{l}
\mathcal{J}^v = tr\left((\bm{W}^v)^T \left( \bm{M}_D-\bm{M}_S  \right)\bm{W}^v \right)\\ ~~~~~~+\frac{1}{2\eta}tr\left((\bm{W}^v)^T\bm{X}^v(\bm{X}^w)^T\bm{W}^w \right)
\end{array}
\end{equation}

And we can update $\alpha_v, v=1,2,\cdots,m$ according to Eq.\ref{eq11}. SM$^2$L iterative  $\bm{\alpha}$ and $\bm{W}^1,\bm{W}^2,\cdots,\bm{W}^m$ alternately. And the iteration stops once the obtained $\bm{W}^1,\bm{W}^2,\cdots,\bm{W}^m$ converge. 

\section{Experiment}

In this section, we conducted various experiments on benchmark datasets. And the experiments have verified that our proposed SM$^2$L can achieve good performances.

\subsection{Datasets and Comparing Methods}

In this section, we introduced the utilized datasets and the comparing methdos in detail. There are 3 datasets utilized in our experiments, including 3Sources \footnote{http://mlg.ucd.ie/datasets/3sources.html}, Cora \footnote{http://lig-membres.imag.fr/grimal/data.html},  WebKB \footnote{http://www.webkb.org}. All these datasets are benchmark multiview datasets. 

3Sources is a benchmark multiview dataset and consists of  3 well-known online news sources: BBC, Reuters and the Guardian. Each source was utilized as one single view and this dataset selected 169 stories as samples.  Cora contains 2708 publications which come from 7 classes. Contents and cites are utilized as 2 views in Cora dataset. WebKB contains 4 subsets of documents over 6 categories. All samples in WebKB consists features from 3 views in one page, including the text on it, the anchor text on the hyperlink pointing to it and the text in its title.

In order to show the excellent performance of SM$^2$L, we compare it with several distance metric learning methods in the experiments, including ITML \cite{davis2007information}, CMM \cite{wang2011semisupervised}, LMNN \cite{weinberger2006distance} and MML \cite{wang2017multi}. ITML, CMM and LMNN are 3 famous single view DML methods. They cannot utilized features from multiple views at the same time. Therefore, we trained them on these views and shown the best performances. Both MML and our proposed SM$^2$L are multiview DML methods.

\subsection{Experiment Results}

In this section, we conducted experiments on the benchmark datasets which has been described above and shown the results carefully. In our experiments, all methods are trained or tested using the same training or testing samples.  

For 3Sources dataset, our experiment randomly select 120 and 80 samples as training ones while the other ones are assigned as testing ones. All DML methods trained their distance metric matrix after the training phase, the experiment adopts 1NN classification on the testing samples. All the DML methods are conducted 10 times and the mean and max classification accuracies are shown as Table \ref{tab1}.

\begin{table}[htbp]
\centering
\caption{The classification accuracies$\left({\%}\right)$ on 3Sources dataset}
\begin{tabular}
{ccccc}
\hline
\raisebox{-1.50ex}[0cm][0cm]{Methods}&
\multicolumn{2}{c}{120} &
\multicolumn{2}{c}{80}  \\
\cline{2-5}
&Mean&Max&Mean&Max\\
\cline{1-5}
ITML&82.13&89.44&76.38&83.23\\

CMM&61.29&66.01&59.33&64.52\\

LMNN&73.97&84.63&70.76&80.95\\

MML&83.32&92.01&78.83&89.32 \\

SM$^2$L&\textbf{85.07}&\textbf{92.71}&\textbf{79.93}&\textbf{90.32} \\
\hline
\end{tabular}
\label{tab1}
\end{table}

It is clearly that SM$^2$L can achive best performances on 3Sources dataset. As single view DML methods, ITML, CMM and LMNN cannot achieve satisfactory performances. Because MML and SM$^2$L are multiview metric learning methods, they are better than those 3 single view ones. Because multiview algorithms take more information into considerations, they are better in most situations.

For Cora dataset,  our experiment randomly select 1900 and 1400 samples as training ones while the other ones are assigned as testing ones. All DML methods trained their distance metric matrix after the training phase, the experiment adopts 1NN classification on the testing samples. All the DML methods are conducted 10 times and the mean and max classification accuracies are shown as Table \ref{tab1}.

\begin{table}[htbp]
	\centering
	\caption{The classification accuracies$\left({\%}\right)$ on Cora dataset}
	\begin{tabular}
		{ccccc}
		\hline
		\raisebox{-1.50ex}[0cm][0cm]{Methods}&
		\multicolumn{2}{c}{1900} &
		\multicolumn{2}{c}{1400}  \\
		\cline{2-5}
		&Mean&Max&Mean&Max\\
		\cline{1-5}
		ITML&60.89&62.01&60.15&61.76\\
		
		CMM&65.55&67.03&62.39&63.44\\
		
		LMNN&66.58&68.14&63.26&64.89\\
		
		MML&67.14&68.32&64.22&66.32 \\
		
		SM$^2$L&\textbf{67.44}&\textbf{68.59}&\textbf{65.23}&\textbf{66.93} \\
		\hline
	\end{tabular}
	\label{tab1}
\end{table}

SM$^2$L can achieve best performances on Cora dataset. Meanwhile, MML is another good distance metric learning for multiview data. Fpr those 3 single view methods, LMNN and CMM are better than ITML.

For 4 subsets of WebKB, we randomly select $70\%$ samples as training ones while the other(about $30\%$) are selected as testing ones. All DML methods trained their distance metric matrix after the training phase, the experiment adopts 1NN classification on the testing samples. All the DML methods are conducted 10 times and the mean and max classification accuracies are shown as Table \ref{tab3}.

\begin{table}[htbp]
	\centering

	\caption{The classification accuracies$\left({\%}\right)$ on WebKB dataset}
	\begin{tabular}
		{ccccccc}
		\hline
		Subsets&&ITML&CMM&LMNN&MML&SM$^2$L\\
	
		\hline
		\multirow{2}{*}{WebKB1}&Mean&84.23&83.10&84.93&85.07&\textbf{86.12}\\
		                       &Max &86.88&88.83&88.12&90.71&\textbf{91.27}\\
		
		\multirow{2}{*}{WebKB2}&Mean&82.04&81.14&81.32&81.93&\textbf{82.58}\\
						       &Max &91.01&88.07&91.15&91.32&\textbf{91.66}\\
		
		\multirow{2}{*}{WebKB3}&Mean&90.55&89.78&90.33&90.52&\textbf{90.92}\\
						       &Max &93.61&94.05&94.11&92.67&\textbf{93.87}\\
		
		\multirow{2}{*}{WebKB4}&Mean&84.63&82.78&82.01&85.65&\textbf{86.54}\\
		                       &Max &88.22&90.54&88.44&91.35&\textbf{92.33}\\
		\hline
	\end{tabular}
	\label{tab3}
\end{table}

It can be found in Table \ref{tab3} that SM$^2$L can achieve best performances in most situations. For 4 subsets of WebKB, because MML and SM$^2$L are multiview methods which can full exploit multiview data, they are better than those 3 single view methods. For those 3 single view methods, LMNN is the best one for WebKB. 

\section{Conclusion}
In this paper, we propose a novel multiview distance metric learning method named Self-weighted Multiview Metric Learning (SM$^2$L). The proposed SM$^2$L utilized maximum margin criterion to maximize the distances between multiview features from different classes while minimize the distances between multiview features from the same ones. Furthermore, SM$^2$L integrates information from all views maximizing the cross correlations between each 2 views. Finally, we show the solving procedure of SM$^2$L in detail. The experiment results have shown the superiority of SM$^2$L.


\bibliographystyle{IEEEbib}
\bibliography{icme2019template}

\end{document}